\documentclass[runningheads]{llncs}
\usepackage[T1]{fontenc}
\usepackage{cite}
\usepackage{graphicx}
\usepackage{multirow}
\usepackage{adjustbox}
\usepackage{amsfonts}
\usepackage{hyperref}
\usepackage{breakurl}
\usepackage{float}
\usepackage[table,xcdraw]{xcolor}
\begin{document}
\title{Towards Automatic Forecasting: Evaluation of Time-Series Forecasting Models for Chickenpox Cases Estimation in Hungary}
\author{Wadie Skaf\orcidID{0000-0002-4298-6694} \and Arzu Tosayeva \and Dániel T. Várkonyi\orcidID{0000-0002-3703-1332}}

\authorrunning{W. Skaf et al.}
\titlerunning{Time-Series Forecasting Models Evaluation for
Chickenpox Case}
\institute{
Telekom Innovation Laboratories, Data Science and Engineering Department (DSED), Faculty of Informatics, Eötvös Loránd University, Pázmány Péter stny. 1/A, 1117, Budapest, Hungary. \\
\email{\{skaf, n0ndni, varkonyid\}@inf.elte.hu}}
\maketitle

\begin{abstract}
Time-Series Forecasting is a powerful data modeling discipline that analyzes historical observations to predict future values of a time-series. It has been utilized in numerous applications, including but not limited to economics, meteorology, and health. In this paper, we use time-series forecasting techniques to model and predict the future incidence of chickenpox. To achieve this, we implement and simulate multiple models and data preprocessing techniques on a Hungary-collected dataset. We demonstrate that the LSTM model outperforms all other models in the vast majority of the experiments in terms of county-level forecasting, whereas the SARIMAX model performs best at the national level. We also demonstrate that the performance of the traditional data preprocessing method is inferior to that of the data preprocessing method that we have proposed.

\end{abstract}
\section{Introduction}
Varicella Zoster Virus (VZV) is a member of the herpes virus family with double-stranded DNA \cite{arvin1996varicella}. This virus causes varicella (chickenpox), a highly contagious pediatric disease often contracted between the ages of 2 and 8 \cite{arvin1996varicella}. Chickenpox is generally a mild disease, although it can develop problems that necessitate hospitalization \cite{bonanni2009varicella, helmuth2015varicella} and, in rare cases, be fatal \cite{EU_Va_control_2015}. Despite the fact that chickenpox is an extremely contagious disease in which over 90\% of unvaccinated persons become infected \cite{breuer_fifer_2011}, and despite the availability of vaccinations \cite{EU_Va_control_2015}, Hungary has no explicit prescription for chickenpox vaccination in its national immunization policy \cite{EU_Va_control_2015}. Given this, and the fact that the reported cases throughout these years form a time-series of values, studies can be conducted to predict the number of future cases in the country, allowing the health system and necessary medications to be prepared. In this paper, we examine the various models of Time-Series Forecasting and perform model evaluation for the chickenpox cases forecasting use case in order to choose the model that achieves the best results at the county- and country-level. Our primary contributions are as follows:
(1) Conducting a comprehensive exploratory data analysis on this relatively new dataset in order to identify underlying patterns. (2) Examining the relationship between chickenpox cases and other variables such as the population. (3) Conducting comprehensive experiments on multiple time-series models and selecting the model that produces the best results for each county and at the national level.

The paper is structured as follows. First, we list and discuss related work, then we formalize and discuss the issue of time-series forecasting, after that, we do exploratory data analysis (EDA) in which we explore the dataset and list its key characteristics. Finally, we detail the experimental setup before reporting and summarizing our major findings.

\section{Related Work}
Time-Series Forecasting research dates back to 1985 \cite{DeGooijer200525YO}, and since then, it has been a constantly expanding research area, especially in the past decade \cite{Alsharef2022}, due to the expansion of data volumes arising from users, industries, and markets, as well as the centrality of forecasting in various applications, such as economic, weather, stock price, business development, and health. As a result, numerous forecasting models have been developed, including ARIMA \cite{p._jenkins_1976}, SARIMA \cite{p._jenkins_1976}, ARIMAX \cite{Cools2008INVESTIGATINGTV}, SARIMAX \cite{Cools2008INVESTIGATINGTV}, N-BEATS \cite{NBEATS_OreshkinCCB20}, DeepAR \cite{DeepAR_SALINAS20201181}, Long Short-Term Memory Neural Network (LSTM) \cite{LINDEMANN2021650}, Gated Recurrent Unit Neural Networks (GRU) \cite{Zhang2017TimeSF}, and Temporal Fusion Transformer (TFT) \cite{lim2021temporal}. These models and others have been utilized in a wide range of use cases, including but not limited to the following: energy and fuels \cite{MUZAFFAR20192922, pr9091617, FENG20211447} where accurate estimates are required to improve power system planning and operation, Finance \cite{Sezer2020FinancialTS, math8091441, Dingli2017FinancialTS}, Environment \cite{LIU2015183, Zhang2020, CHEN2018681}, Industry \cite{WANG2019144, Rashid2019TimesseriesDA, HUANG2019437}, and Health \cite{10.1007/978-981-10-4361-1_138, Hoppe2017DeepLF, Sarafrazi2019CrackingT}. In this paper, we contribute to the usage of time-series forecasting in the Health domain by predicting the number of cases of chickenpox in Hungary.

\section{Time-Series Forecasting Problem Definition}
Time-series forecasting problem can be formalized as follows: given a univariate time-series, which represents a sequence of values $X = (\mathbf{x}_{1}, \mathbf{x}_{2}, \dots, \mathbf{x}_{t})$ forecasting is the process of predicting the value of future observations of a time-series ($\mathbf{x}_{t+1}, \mathbf{x}_{t+2}, \dots, \mathbf{x}_{t+h})$  based on historical data,  where $\mathbf{x}_{i} \in \mathbb{R}$ $(i\in [1,\dots,t+h])$ is the value of $X$ at time $i$, $t$ is the length of $X$, and $h$ is the forecasting horizon.

\section{Exploratory Data Analysis (EDA)}
\begin{figure}[]
    \centering
    \includegraphics[width=0.8\textwidth]{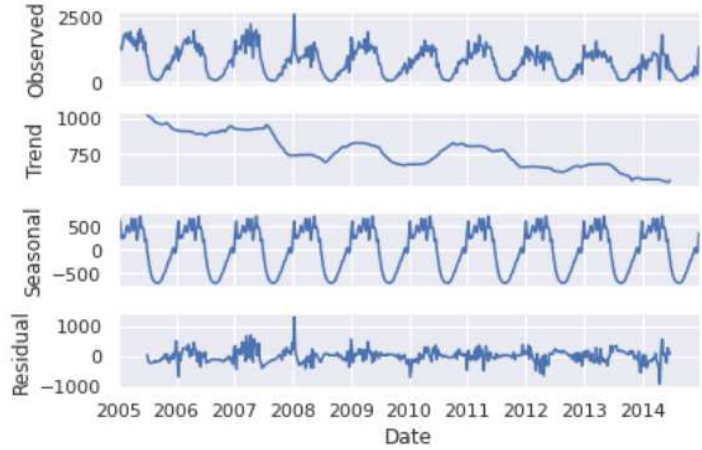}
    \caption{Country-level Time-Series Data Decomposition}
    \label{fig:country_level_decomposition}
\end{figure}

\begin{figure*}[t]
    \centering
    \includegraphics[width=\textwidth]{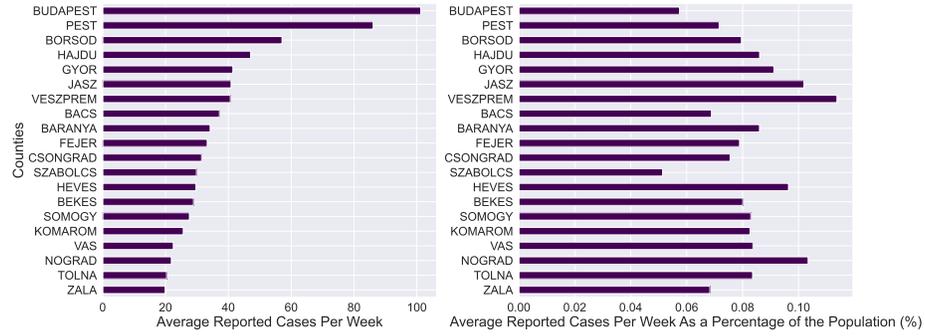}
    \caption{The average number of weekly reported cases per county}
    \label{fig:avg_cases_county}
\end{figure*}

\begin{figure*}[t]
    \centering
    \includegraphics[width=\linewidth]{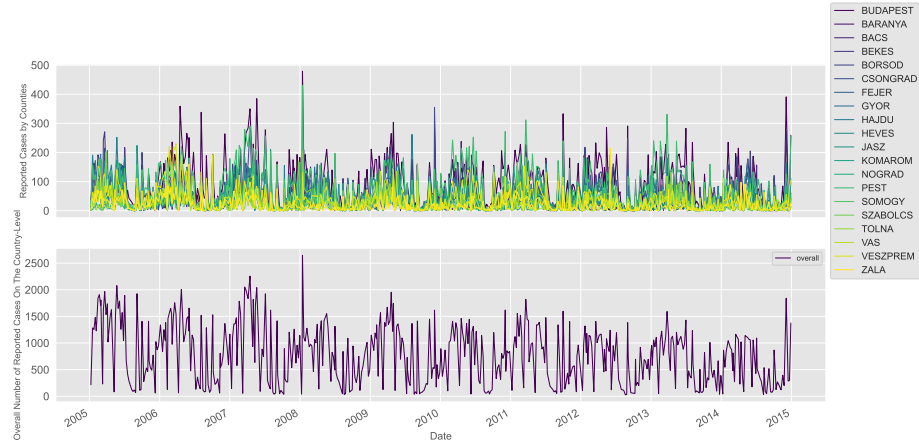}
    \caption{Chickenpox weekly cases in Hungary reported between 2005 and 2015 broken by counties}
    \label{fig:all_counties_data}
\end{figure*}

The dataset used in this paper was made available by Rozemberczki, B. et al \cite{rozemberczki2021chickenpox}. This dataset consists of county-level time series depicting the weekly number of chickenpox cases reported by general practitioners in Hungary from 2005 to 2015, subdivided into 20 vertices: Budapest, Pest, Borsod, Hajdu, Gyor, Jasz, Veszprem, Bacs, Baranya, Fejer, Csongrad, Szabolcs, Heves, Bekes, Somogy, Komarom, Vas, Nograd, Torna, and Zala. The main charactericts of the data are as following:

\begin{enumerate}
    \item As can be seen in Figure \ref{fig:avg_cases_county}, the city of Budapest, Hungary's capital, has the highest average number of reported cases each week by a significant margin compared to other counties. This is primarily due to the difference in population, and consequently, if we calculate the average number of reported cases as a percentage of the population, we can deduce that Veszprem has the highest ratio.
    \item In Figure \ref{fig:all_counties_data}, seasonality is evident, with the greatest number of cases occurring during the winter months and the smallest number occurring between the summer and fall seasons. This is also apparent by decomposing the country-level time series (Figure \ref{fig:country_level_decomposition}).
    \item A downward trend can be noticed in the data (Figure \ref{fig:country_level_decomposition}).
\end{enumerate}

\section{Experimental Setup}
\subsection{Data Splitting}
In all of our experiments, we split the data so that 80 percent of the data is used for training and 20 percent is used for testing; accordingly, the last 20 percent of each series' values are used for testing.
\subsection{Data Normalization}
\label{sec:data_preprocessing}
We experimented two methods of data normalization: 
\begin{enumerate}
    \item Method 1: We performed traditional data normalization so data would be within the range $[-1,1]$.
    \item Method 2: We performed normalization by converting each data sample to a percentage of the population at the time when cases were reported as shown in Equation \ref{eq:pop_normalization}
    \begin{equation}
    \label{eq:pop_normalization}
        \mathbf{x}^{\prime}_{c, d} = \frac{\mathbf{x}_{c, d}}{\mathbf{p}_{c, d}} \times 100
    \end{equation}
    Where $\mathbf{s}_{c, d}$ denotes the reported cases in county $\mathbf{c}$ on date $\mathbf{d}$, and $\mathbf{p}_{c, d}$ denotes the population of county $\mathbf{c}$ on date $\mathbf{d}$.

\end{enumerate}
\subsection{Models}
We conducted experiments on the following models: ARIMA \cite{p._jenkins_1976}, SARIMA \cite{p._jenkins_1976}, SARIMAX \cite{Cools2008INVESTIGATINGTV}, N-BEATS \cite{NBEATS_OreshkinCCB20}, DeepAR \cite{DeepAR_SALINAS20201181}, Long Short-Term Memory Neural Network (LSTM) \cite{LINDEMANN2021650}, Gated Recurrent Unit Neural Networks (GRU) \cite{Zhang2017TimeSF}, and Temporal Fusion Transformer (TFT) \cite{lim2021temporal}.
Throughout the experiments, each model was trained for 200 epochs using the Adam optimizer and a learning rate of $ \alpha = 0.01$.

\subsection{Evaluation Metrics}
To evaluate the performance of the models, we calculated the Root Mean Square Error (RMSE) using the equation:
\begin{equation}
\label{eq:RMSE}
    RMSE= \sqrt{\frac{1}{t}\sum_{i=1}^{t}[\hat x_{i}-x_{i}]^{2}}
\end{equation}
Where $\hat x_{i}$ denotes the $i^{th}$ predicted value and $x_{i}$ denotes the $i^{th}$ original (observed) value.

\section{Benchmarking Results}

The results of the experiments are summarized in table \ref{tab:benchmarking results}, which contains a collection of the results for each model, separated according to the county and the normalization methods. The column labeled "loss 1" refers to method 1, and the column labeled "loss 2" refers to method 2, both of which are described in section \ref{sec:data_preprocessing}.

As can be seen in table \ref{tab:benchmarking results}, regarding forecasting of the individual county, the LSTM model performs better than any of the other models in a vast majority of the counties:  Budapest, Bekes, Heves, Szabolcs, Veszprem, Baranya, Borsod, Jasz, Pest, Tolna, Zala, Bacs, Csongrad, Hajdu, Komarom, Somogy, and Vas with the RMSE loss values of 0.03, 0.03, 0.04, 0.04, 0.05, 0.03, 0.05, 0.03, 0.04, 0.05, 0.04, 0.04, 0.04, 0.03, 0.05, 0.03, 0.06 respectively and the GRU Model performed better in Fejer, Nograd, and Gyor with RMSE loss values of 0.04, 0.02, and 0.03, respectively, whereas when it came to forecasting on the national level, SARIMAX achieved the best results when with an RMSE loss value of 0.02. The main reason the LSTM model outperformed other models in most cases was due to its ability to do long-term memorization more than the other models, and as can be seen in Figure \ref{fig:all_counties_data}, the vast majority of the series does not have a consistent pattern and would benefit from this long-term memorization; this also explains why SARIMAX performed better on the country-level series (Figures \ref{fig:all_counties_data} and \ref{fig:country_level_decomposition}), where all the series were summed, resulting in a more consistent pattern that does not rely heavily on long-term memorization ability.

\begin{figure}[H]
    \centering
    \includegraphics[width=0.65\textwidth]{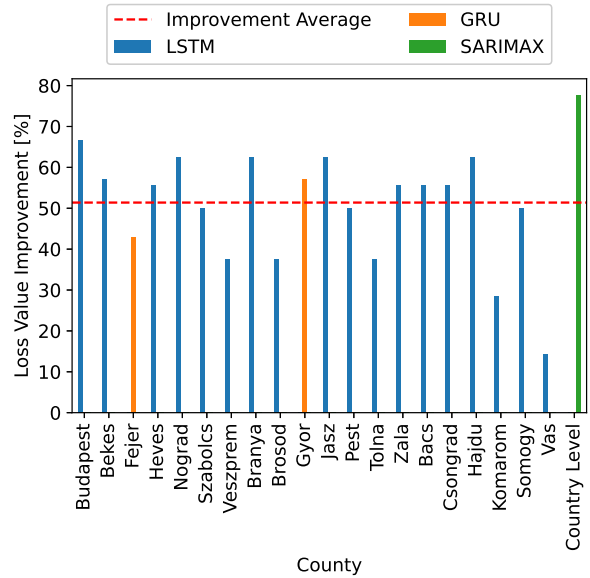}
    \caption{Improvements in RMSE Loss values after applying the Normalization Method 2 in comparison to Normalization Method 1}
    \label{fig:loss_improvement}
\end{figure}

In addition, the normalization approach that we have proposed outperformed the traditional normalization approach (Method 1) in each and every (model, county) pair experiment, achieving a significant improvement in terms of the RMSE loss value. By calculating the loss value improvement after applying Method 2 in comparison to Method 1 for the best performing model for each county (as highlighted in Table \ref{tab:benchmarking results}), we can see that the SARIMAX model for the country-level forecasting has the highest gain with a $77.78 \%$ improvement, while the LSTM model for Vas county has the lowest gain with a $14.29 \%$ improvement, and the average improvement across all models is $51.39\%$. (Figure \ref{fig:loss_improvement}).

\section{Conclusion}
In this paper, we presented, discussed, and highlighted the results of a series of experiments that we conducted out on the chickenpox cases dataset. The purpose of these experiments was to evaluate time-series forecasting models for use in predicting the number of chickenpox cases in Hungary at the county and national levels. We demonstrated that the LSTM model performed better than other models for the majority of county-level forecasting except in the cases of Fejer, Nograd, and Gyor counties, while the SARIMAX model produced the most accurate results at the country-level. In addition, we proposed a custom data preprocessing method for this dataset by dividing the proportion of cases by the population size, and demonstrated that this method outperformed conventional normalization in terms of achieving lower RMSE Loss values.
\begin{table}[]

\centering
\caption{Benchmarking Results}
\label{tab:benchmarking results}
\begin{adjustbox}{width=\textwidth}
\begin{tabular}{|c|c|c|c|c|c|c|c|c|c|c|c|}
\hline
\multicolumn{1}{|c|}{County} &
  Model &
  \multicolumn{1}{c|}{\begin{tabular}[c]{@{}l@{}}Loss1\end{tabular}} &
  \multicolumn{1}{c|}{\begin{tabular}[c]{@{}l@{}}Loss2\end{tabular}} &
  \multicolumn{1}{c|}{County} &
  Model &
  \multicolumn{1}{c|}{\begin{tabular}[c]{@{}l@{}}Loss1\end{tabular}} &
  \multicolumn{1}{c|}{\begin{tabular}[c]{@{}l@{}}Loss 2\end{tabular}} &
  \multicolumn{1}{c|}{County} &
  Model &
  \multicolumn{1}{c|}{\begin{tabular}[c]{@{}l@{}}Loss 1\end{tabular}} &
  \multicolumn{1}{c|}{\begin{tabular}[c]{@{}l@{}}Loss 2\end{tabular}} \\ \hline
 &
  Sarimax &
  0.11 &
  0.04 &
   &
  Sarimax &
  0.12 &
  0.04 &
   &
  Sarimax &
  0.12 &
  0.07 \\ \cline{2-4} \cline{6-8} \cline{10-12} 
 &
  Arima &
  0.35 &
  0.21 &
   &
  Arima &
  0.33 &
  0.18 &
   &
  Arima &
  0.35 &
  0.21 \\ \cline{2-4} \cline{6-8} \cline{10-12} 
 &
  Sarima &
  0.24 &
  0.14 &
   &
  Sarima &
  0.28 &
  0.15 &
   &
  Sarima &
  0.22 &
  0.12 \\ \cline{2-4} \cline{6-8} \cline{10-12} 
 &
  \textbf{LSTM} &
  0.09 &
  \textbf{0.03} &
   &
  \textbf{LSTM} &
  0.08 &
  \textbf{0.03} &
   &
  \textbf{LSTM} &
  0.09 &
  \textbf{0.04} \\ \cline{2-4} \cline{6-8} \cline{10-12} 
 &
  GRU &
  0.13 &
  0.07 &
   &
  GRU &
  0.12 &
  0.06 &
   &
  GRU &
  0.12 &
  0.08 \\ \cline{2-4} \cline{6-8} \cline{10-12} 
 &
  N-BEATS &
  0.30 &
  0.20 &
   &
  N-BEATS &
  0.22 &
  0.05 &
   &
  N-BEATS &
  0.30 &
  0.05 \\ \cline{2-4} \cline{6-8} \cline{10-12} 
 &
  DeepAR &
  0.17 &
  0.09 &
   &
  DeepAR &
  0.15 &
  0.05 &
   &
  DeepAR &
  0.14 &
  0.08 \\ \cline{2-4} \cline{6-8} \cline{10-12} 
\multirow{-8}{*}{Budapest} &
  TFT &
  0.34 &
  0.11 &
  \multirow{-8}{*}{Baranya} &
  TFT &
  0.23 &
  0.09 &
  \multirow{-8}{*}{Bacs} &
  TFT &
  0.22 &
  0.12 \\ \hline
 &
  Sarimax &
  0.13 &
  0.05 &
   &
  Sarimax &
  0.12 &
  0.08 &
   &
  Sarimax &
  0.13 &
  0.07 \\ \cline{2-4} \cline{6-8} \cline{10-12} 
 &
  Arima &
  0.35 &
  0.20 &
   &
  Arima &
  0.27 &
  0.13 &
   &
  Arima &
  0.34 &
  0.21 \\ \cline{2-4} \cline{6-8} \cline{10-12} 
 &
  Sarima &
  0.19 &
  0.11 &
   &
  Sarima &
  0.15 &
  0.11 &
   &
  Sarima &
  0.19 &
  0.11 \\ \cline{2-4} \cline{6-8} \cline{10-12} 
 &
  \textbf{LSTM} &
  0.07 &
  \textbf{0.03} &
   &
  \textbf{LSTM} &
  0.08 &
  \textbf{0.05} &
   &
  \textbf{LSTM} &
  0.09 &
  \textbf{0.04} \\ \cline{2-4} \cline{6-8} \cline{10-12} 
 &
  GRU &
  0.10 &
  0.08 &
   &
  GRU &
  0.09 &
  0.06 &
   &
  GRU &
  0.09 &
  0.07 \\ \cline{2-4} \cline{6-8} \cline{10-12} 
 &
  N-BEATS &
  0.25 &
  0.04 &
   &
  N-BEATS &
  0.20 &
  0.07 &
   &
  N-BEATS &
  0.32 &
  0.08 \\ \cline{2-4} \cline{6-8} \cline{10-12} 
 &
  DeepAR &
  0.16 &
  0.09 &
   &
  DeepAR &
  0.18 &
  0.08 &
   &
  Deep-AR &
  0.14 &
  0.07 \\ \cline{2-4} \cline{6-8} \cline{10-12} 
\multirow{-8}{*}{Bekes} &
  TFT &
  0.22 &
  0.06 &
  \multirow{-8}{*}{Borsod} &
  TFT &
  0.32 &
  0.08 &
  \multirow{-8}{*}{Csongrad} &
  TFT &
  0.25 &
  0.09 \\ \hline
 &
  Sarimax &
  0.11 &
  0.06 &
   &
  Sarimax &
  0.12 &
  0.05 &
   &
  Sarimax &
  0.12 &
  0.07 \\ \cline{2-4} \cline{6-8} \cline{10-12} 
 &
  Arima &
  0.35 &
  0.21 &
   &
  Arima &
  0.33 &
  0.18 &
   &
  Arima &
  0.34 &
  0.21 \\ \cline{2-4} \cline{6-8} \cline{10-12} 
 &
  Sarima &
  0.24 &
  0.14 &
   &
  Sarima &
  0.28 &
  0.15 &
   &
  Sarima &
  0.19 &
  0.11 \\ \cline{2-4} \cline{6-8} \cline{10-12} 
 &
  LSTM &
  0.09 &
  0.06 &
   &
  LSTM &
  0.07 &
  0.04 &
   &
  \textbf{LSTM} &
  0.08 &
  \textbf{0.03} \\ \cline{2-4} \cline{6-8} \cline{10-12} 
 &
  \textbf{GRU} &
  0.07 &
  \textbf{0.04} &
   &
  \textbf{GRU} &
  0.07 &
  \textbf{0.03} &
   &
  GRU &
  0.12 &
  0.08 \\ \cline{2-4} \cline{6-8} \cline{10-12} 
 &
  N-BEATS &
  0.23 &
  0.06 &
   &
  N-BEATS &
  0.30 &
  0.04 &
   &
  N-BEATS &
  0.23 &
  0.06 \\ \cline{2-4} \cline{6-8} \cline{10-12} 
 &
  DeepAR &
  0.15 &
  0.08 &
   &
  DeepAR &
  0.18 &
  0.07 &
   &
  DeepAR &
  0.15 &
  0.07 \\ \cline{2-4} \cline{6-8} \cline{10-12} 
\multirow{-8}{*}{Fejer} &
  TFT &
  0.25 &
  0.12 &
  \multirow{-8}{*}{Gyor} &
  TFT &
  0.26 &
  0.09 &
  \multirow{-8}{*}{Hajdu} &
  TFT &
  0.24 &
  0.12 \\ \hline
 &
  Sarimax &
  0.12 &
  0.05 &
   &
  Sarimax &
  0.10 &
  0.04 &
   &
  Sarimax &
  0.11 &
  0.06 \\ \cline{2-4} \cline{6-8} \cline{10-12} 
 &
  Arima &
  0.36 &
  0.18 &
   &
  Arima &
  0.35 &
  0.22 &
   &
  Arima &
  0.33 &
  0.13 \\ \cline{2-4} \cline{6-8} \cline{10-12} 
 &
  Sarima &
  0.26 &
  0.12 &
   &
  Sarima &
  0.24 &
  0.11 &
   &
  Sarima &
  0.21 &
  0.12 \\ \cline{2-4} \cline{6-8} \cline{10-12} 
 &
  \textbf{LSTM} &
  0.09 &
  \textbf{0.04} &
   &
  \textbf{LSTM} &
  0.08 &
  \textbf{0.03} &
   &
  \textbf{LSTM} &
  0.07 &
  \textbf{0.05} \\ \cline{2-4} \cline{6-8} \cline{10-12} 
 &
  GRU &
  0.12 &
  0.07 &
   &
  GRU &
  0.09 &
  0.05 &
   &
  GRU &
  0.11 &
  0.08 \\ \cline{2-4} \cline{6-8} \cline{10-12} 
 &
  N-BEATS &
  0.11 &
  0.05 &
   &
  N-BEATS &
  0.22 &
  0.06 &
   &
  N-BEATS &
  0.33 &
  0.07 \\ \cline{2-4} \cline{6-8} \cline{10-12} 
 &
  DeepAR &
  0.13 &
  0.06 &
   &
  DeepAR &
  0.14 &
  0.07 &
   &
  DeepAR &
  0.16 &
  0.07 \\ \cline{2-4} \cline{6-8} \cline{10-12} 
\multirow{-8}{*}{Heves} &
  TFT &
  0.22 &
  0.11 &
  \multirow{-8}{*}{Jasz} &
  TFT &
  0.33 &
  0.13 &
  \multirow{-8}{*}{Komarom} &
  TFT &
  0.34 &
  0.09 \\ \hline
 &
  Sarimax &
  0.12 &
  0.06 &
   &
  Sarimax &
  0.13 &
  0.05 &
   &
  Sarimax &
  0.12 &
  0.04 \\ \cline{2-4} \cline{6-8} \cline{10-12} 
 &
  Arima &
  0.33 &
  0.18 &
   &
  Arima &
  0.33 &
  0.13 &
   &
  Arima &
  0.36 &
  0.18 \\ \cline{2-4} \cline{6-8} \cline{10-12} 
 &
  Sarima &
  0.28 &
  0.15 &
   &
  Sarima &
  0.21 &
  0.12 &
   &
  Sarima &
  0.26 &
  0.12 \\ \cline{2-4} \cline{6-8} \cline{10-12} 
 &
  LSTM &
  0.08 &
  0.03 &
   &
  \textbf{LSTM} &
  0.08 &
  \textbf{0.04} &
   &
  \textbf{LSTM} &
  0.06 &
  \textbf{0.03} \\ \cline{2-4} \cline{6-8} \cline{10-12} 
 &
  \textbf{GRU} &
  0.09 &
  \textbf{0.02} &
   &
  GRU &
  0.11 &
  0.06 &
   &
  GRU &
  0.12 &
  0.07 \\ \cline{2-4} \cline{6-8} \cline{10-12} 
 &
  N-BEATS &
  0.24 &
  0.04 &
   &
  N-BEATS &
  0.33 &
  0.05 &
   &
  N-BEATS &
  0.32 &
  0.06 \\ \cline{2-4} \cline{6-8} \cline{10-12} 
 &
  DeepAR &
  0.14 &
  0.07 &
   &
  DeepAR &
  0.14 &
  0.07 &
   &
  DeepAR &
  0.13 &
  0.06 \\ \cline{2-4} \cline{6-8} \cline{10-12} 
\multirow{-8}{*}{Nograd} &
  TFT &
  0.24 &
  0.11 &
  \multirow{-8}{*}{Pest} &
  TFT &
  0.23 &
  0.11 &
  \multirow{-8}{*}{Somogy} &
  TFT &
  0.26 &
  0.12 \\ \hline
 &
  Sarimax &
  0.14 &
  0.08 &
   &
  Sarimax &
  0.11 &
  0.06 &
   &
  Sarimax &
  0.12 &
  0.07 \\ \cline{2-4} \cline{6-8} \cline{10-12} 
 &
  Arima &
  0.38 &
  0.22 &
   &
  Arima &
  0.34 &
  0.21 &
   &
  Arima &
  0.33 &
  0.13 \\ \cline{2-4} \cline{6-8} \cline{10-12} 
 &
  Sarima &
  0.26 &
  0.13 &
   &
  Sarima &
  0.23 &
  0.11 &
   &
  Sarima &
  0.21 &
  0.12 \\ \cline{2-4} \cline{6-8} \cline{10-12} 
 &
  \textbf{LSTM} &
  0.08 &
  \textbf{0.04} &
   &
  \textbf{LSTM} &
  0.08 &
  \textbf{0.05} &
   &
  \textbf{LSTM} &
  0.07 &
  \textbf{0.06} \\ \cline{2-4} \cline{6-8} \cline{10-12} 
 &
  GRU &
  0.09 &
  0.07 &
   &
  GRU &
  0.11 &
  0.07 &
   &
  GRU &
  0.12 &
  0.08 \\ \cline{2-4} \cline{6-8} \cline{10-12} 
 &
  N-BEATS &
  0.23 &
  0.06 &
   &
  N-BEATS &
  0.33 &
  0.06 &
   &
  N-BEATS &
  0.23 &
  0.07 \\ \cline{2-4} \cline{6-8} \cline{10-12} 
 &
  DeepAR &
  0.15 &
  0.09 &
   &
  DeepAR &
  0.19 &
  0.08 &
   &
  DeepAR &
  0.18 &
  0.08 \\ \cline{2-4} \cline{6-8} \cline{10-12} 
\multirow{-8}{*}{Szabolcs} &
  TFT &
  0.33 &
  0.09 &
  \multirow{-8}{*}{Tolna} &
  TFT &
  0.33 &
  0.12 &
  \multirow{-8}{*}{Vas} &
  TFT &
  0.22 &
  0.09 \\ \hline
\multicolumn{1}{|l|}{} &
  Sarimax &
  0.12 &
  0.06 &
   &
  Sarimax &
  0.11 &
  0.07 &
   &
  \textbf{Sarimax} &
  0.09 &
  \textbf{0.02} \\ \cline{2-4} \cline{6-8} \cline{10-12} 
\multicolumn{1}{|l|}{} &
  Arima &
  0.35 &
  0.21 &
   &
  Arima &
  0.33 &
  0.23 &
   &
  Arima &
  0.31 &
  0.11 \\ \cline{2-4} \cline{6-8} \cline{10-12} 
\multicolumn{1}{|l|}{} &
  Sarima &
  0.22 &
  0.12 &
   &
  Sarima &
  0.35 &
  0.21 &
   &
  Sarima &
  0.25 &
  0.14 \\ \cline{2-4} \cline{6-8} \cline{10-12} 
\multicolumn{1}{|l|}{} &
  \textbf{LSTM} &
  0.08 &
  \textbf{0.05} &
   &
  \textbf{LSTM} &
  0.09 &
  \textbf{0.04} &
   &
  LSTM &
  0.09 &
  0.07 \\ \cline{2-4} \cline{6-8} \cline{10-12} 
\multicolumn{1}{|l|}{} &
  GRU &
  0.12 &
  0.07 &
   &
  GRU &
  0.13 &
  0.08 &
   &
  GRU &
  0.08 &
  0.06 \\ \cline{2-4} \cline{6-8} \cline{10-12} 
\multicolumn{1}{|l|}{} &
  N-BEATS &
  0.22 &
  0.06 &
   &
  N-BEATS &
  0.20 &
  0.05 &
   &
  N-BEATS &
  0.23 &
  0.03 \\ \cline{2-4} \cline{6-8} \cline{10-12} 
\multicolumn{1}{|l|}{} &
  DeepAR &
  0.16 &
  0.07 &
   &
  DeepAR &
  0.15 &
  0.07 &
   &
  DeepAR &
  0.19 &
  0.08 \\ \cline{2-4} \cline{6-8} \cline{10-12} 
\multicolumn{1}{|l|}{\multirow{-8}{*}{Veszprem}} &
  TFT &
  0.34 &
  0.12 &
  \multirow{-8}{*}{Zala} &
  TFT &
  0.30 &
  0.11 &
  \multirow{-8}{*}{\shortstack{Country \\ Level}} &
  TFT &
  0.33 &
  0.12 \\ \hline
\end{tabular}
\end{adjustbox}%


\end{table}
\newpage
\bibliographystyle{splncs04}
%
\bibliography{citations}

\begin{thebibliography}{10}
\providecommand{\url}[1]{\texttt{#1}}
\providecommand{\urlprefix}{URL }
\providecommand{\doi}[1]{https://doi.org/#1}

\bibitem{EU_Va_control_2015}
Public health guidance on varicella vaccination in the european union (Feb
  2015), available at
  \url{https://www.ecdc.europa.eu/en/publications-data/public-health-guidance-varicella-vaccination-european-union}

\bibitem{Alsharef2022}
Alsharef, A., Aggarwal, K., {Sonia}, Kumar, M., Mishra, A.: Review of ml and
  automl solutions to forecast time-series data. Archives of Computational
  Methods in Engineering  (Jun 2022). \doi{10.1007/s11831-022-09765-0},
  \url{https://doi.org/10.1007/s11831-022-09765-0}

\bibitem{arvin1996varicella}
Arvin, A.M.: Varicella-zoster virus. Clinical microbiology reviews
  \textbf{9}(3),  361--381 (1996)

\bibitem{bonanni2009varicella}
Bonanni, P., Breuer, J., Gershon, A., Gershon, M., Hryniewicz, W.,
  Papaevangelou, V., Rentier, B., R{\"u}mke, H., Sadzot-Delvaux, C., Senterre,
  J., et~al.: Varicella vaccination in europe--taking the practical approach.
  BMC medicine  \textbf{7}(1),  1--12 (2009)

\bibitem{breuer_fifer_2011}
Breuer, J., Fifer, H.: Chickenpox (Apr 2011),
  \url{https://www.ncbi.nlm.nih.gov/pmc/articles/PMC3275319/}

\bibitem{10.1007/978-981-10-4361-1_138}
Bui, C., Pham, N., Vo, A., Tran, A., Nguyen, A., Le, T.: Time series
  forecasting for healthcare diagnosis and prognostics with the focus on
  cardiovascular diseases. In: Vo~Van, T., Nguyen~Le, T.A., Nguyen~Duc, T.
  (eds.) 6th International Conference on the Development of Biomedical
  Engineering in Vietnam (BME6). pp. 809--818. Springer Singapore, Singapore
  (2018)

\bibitem{CHEN2018681}
Chen, J., Zeng, G.Q., Zhou, W., Du, W., Lu, K.D.: Wind speed forecasting using
  nonlinear-learning ensemble of deep learning time series prediction and
  extremal optimization. Energy Conversion and Management  \textbf{165},
  681--695 (2018). \doi{https://doi.org/10.1016/j.enconman.2018.03.098},
  \url{https://www.sciencedirect.com/science/article/pii/S0196890418303261}

\bibitem{Cools2008INVESTIGATINGTV}
Cools, M., Moons, E., Wets, G.: Investigating the variability in daily traffic
  counts using arimax and sarima(x) models: Assessing the impact of holidays on
  two divergent site locations (2008)

\bibitem{Dingli2017FinancialTS}
Dingli, A., Fournier, K.S.: Financial time series forecasting – a deep
  learning approach. International Journal of Machine Learning and Computing
  \textbf{7},  118--122 (2017)

\bibitem{FENG20211447}
Feng, Q., Qian, S.: Research on power load forecasting model of economic
  development zone based on neural network. Energy Reports  \textbf{7},
  1447--1452 (2021). \doi{https://doi.org/10.1016/j.egyr.2021.09.098},
  \url{https://www.sciencedirect.com/science/article/pii/S2352484721009045},
  2021 International Conference on Energy Engineering and Power Systems

\bibitem{DeGooijer200525YO}
Gooijer, J.G.D., Hyndman, R.J.: 25 years of iif time series forecasting: A
  selective review. Econometrics eJournal  (2005)

\bibitem{helmuth2015varicella}
Helmuth, I.G., Poulsen, A., Suppli, C.H., M{\o}lbak, K.: Varicella in
  europe—a review of the epidemiology and experience with vaccination.
  Vaccine  \textbf{33}(21),  2406--2413 (2015)

\bibitem{Hoppe2017DeepLF}
Hoppe, E., K{\"o}rzd{\"o}rfer, G., W{\"u}rfl, T., Wetzl, J., Lugauer, F.,
  Pfeuffer, J., Maier, A.K.: Deep learning for magnetic resonance
  fingerprinting: A new approach for predicting quantitative parameter values
  from time series. Studies in health technology and informatics  \textbf{243},
   202--206 (2017)

\bibitem{HUANG2019437}
Huang, X., Zanni-Merk, C., Crémilleux, B.: Enhancing deep learning with
  semantics: an application to manufacturing time series analysis. Procedia
  Computer Science  \textbf{159},  437--446 (2019).
  \doi{https://doi.org/10.1016/j.procs.2019.09.198},
  \url{https://www.sciencedirect.com/science/article/pii/S1877050919313808},
  knowledge-Based and Intelligent Information \& Engineering Systems:
  Proceedings of the 23rd International Conference KES2019

\bibitem{lim2021temporal}
Lim, B., Ar{\i}k, S.{\"O}., Loeff, N., Pfister, T.: Temporal fusion
  transformers for interpretable multi-horizon time series forecasting.
  International Journal of Forecasting  \textbf{37}(4),  1748--1764 (2021)

\bibitem{LINDEMANN2021650}
Lindemann, B., Müller, T., Vietz, H., Jazdi, N., Weyrich, M.: A survey on long
  short-term memory networks for time series prediction. Procedia CIRP
  \textbf{99},  650--655 (2021).
  \doi{https://doi.org/10.1016/j.procir.2021.03.088},
  \url{https://www.sciencedirect.com/science/article/pii/S2212827121003796},
  14th CIRP Conference on Intelligent Computation in Manufacturing Engineering,
  15-17 July 2020

\bibitem{LIU2015183}
Liu, H., qi~Tian, H., feng Liang, X., fei Li, Y.: Wind speed forecasting
  approach using secondary decomposition algorithm and elman neural networks.
  Applied Energy  \textbf{157},  183--194 (2015).
  \doi{https://doi.org/10.1016/j.apenergy.2015.08.014},
  \url{https://www.sciencedirect.com/science/article/pii/S0306261915009393}

\bibitem{MUZAFFAR20192922}
Muzaffar, S., Afshari, A.: Short-term load forecasts using lstm networks.
  Energy Procedia  \textbf{158},  2922--2927 (2019).
  \doi{https://doi.org/10.1016/j.egypro.2019.01.952},
  \url{https://www.sciencedirect.com/science/article/pii/S1876610219310008},
  innovative Solutions for Energy Transitions

\bibitem{NBEATS_OreshkinCCB20}
Oreshkin, B.N., Carpov, D., Chapados, N., Bengio, Y.: {N-BEATS:} neural basis
  expansion analysis for interpretable time series forecasting. In: 8th
  International Conference on Learning Representations, {ICLR} 2020, Addis
  Ababa, Ethiopia, April 26-30, 2020. OpenReview.net (2020),
  \url{https://openreview.net/forum?id=r1ecqn4YwB}

\bibitem{p._jenkins_1976}
P., B.G.E., Jenkins, G.M.: Time Series Analysis Forescasting and Control.
  Holden-Day (1976)

\bibitem{pr9091617}
Qian, K., Wang, X., Yuan, Y.: Research on regional short-term power load
  forecasting model and case analysis. Processes  \textbf{9}(9) (2021).
  \doi{10.3390/pr9091617}, \url{https://www.mdpi.com/2227-9717/9/9/1617}

\bibitem{Rashid2019TimesseriesDA}
Rashid, K.M., Louis, J.: Times-series data augmentation and deep learning for
  construction equipment activity recognition. Adv. Eng. Informatics
  \textbf{42} (2019)

\bibitem{rozemberczki2021chickenpox}
Rozemberczki, B., Scherer, P., Kiss, O., Sarkar, R., Ferenci, T.: Chickenpox
  cases in hungary: a benchmark dataset for spatiotemporal signal processing
  with graph neural networks. arXiv preprint arXiv:2102.08100  (2021)

\bibitem{DeepAR_SALINAS20201181}
Salinas, D., Flunkert, V., Gasthaus, J., Januschowski, T.: Deepar:
  Probabilistic forecasting with autoregressive recurrent networks.
  International Journal of Forecasting  \textbf{36}(3),  1181--1191 (2020).
  \doi{https://doi.org/10.1016/j.ijforecast.2019.07.001},
  \url{https://www.sciencedirect.com/science/article/pii/S0169207019301888}

\bibitem{Sarafrazi2019CrackingT}
Sarafrazi, S., Choudhari, R.S., Mehta, C., Mehta, H.K., Japalaghi, O.K., Han,
  J., Mehta, K.A., Han, H.W., Francis-Lyon, P.A.: Cracking the “sepsis”
  code: Assessing time series nature of ehr data, and using deep learning for
  early sepsis prediction. 2019 Computing in Cardiology (CinC) pp. Page 1--Page
  4 (2019)

\bibitem{Sezer2020FinancialTS}
Sezer, O.B., Gudelek, M.U., Ozbayoglu, A.M.: Financial time series forecasting
  with deep learning : A systematic literature review: 2005-2019. ArXiv
  \textbf{abs/1911.13288} (2020)

\bibitem{math8091441}
Shahi, T.B., Shrestha, A., Neupane, A., Guo, W.: Stock price forecasting with
  deep learning: A comparative study. Mathematics  \textbf{8}(9) (2020).
  \doi{10.3390/math8091441}, \url{https://www.mdpi.com/2227-7390/8/9/1441}

\bibitem{WANG2019144}
Wang, Y., Zhang, D., Liu, Y., Dai, B., Lee, L.H.: Enhancing transportation
  systems via deep learning: A survey. Transportation Research Part C: Emerging
  Technologies  \textbf{99},  144--163 (2019).
  \doi{https://doi.org/10.1016/j.trc.2018.12.004},
  \url{https://www.sciencedirect.com/science/article/pii/S0968090X18304108}

\bibitem{Zhang2017TimeSF}
Zhang, X., Furao, S., Zhao, J., Yang, G.: Time series forecasting using gru
  neural network with multi-lag after decomposition. In: ICONIP (2017)

\bibitem{Zhang2020}
Zhang, Y., Pan, G.: A hybrid prediction model for forecasting wind energy
  resources. Environmental Science and Pollution Research  \textbf{27}(16),
  19428--19446 (Jun 2020). \doi{10.1007/s11356-020-08452-6},
  \url{https://doi.org/10.1007/s11356-020-08452-6}

\end{thebibliography}
\end{document}